\documentclass{article} 
\usepackage{groundgui,times}

\usepackage{amsmath,amsfonts,bm}

\def\eqref#1{equation~\ref{#1}}

\def\1{\bm{1}}

\DeclareMathAlphabet{\mathsfit}{\encodingdefault}{\sfdefault}{m}{sl}
\SetMathAlphabet{\mathsfit}{bold}{\encodingdefault}{\sfdefault}{bx}{n}

\usepackage{hyperref}
\usepackage{url}
\usepackage{booktabs}
\usepackage[table]{xcolor}
\usepackage{multirow}
\definecolor{avgcol}{RGB}{223,240,216}   
\definecolor{secrow}{RGB}{236,236, 236}   
\definecolor{citecol}{RGB}{126,187,79}   
\usepackage{graphicx}
\usepackage{xcolor}
\usepackage{authblk}
\definecolor{anchcol}{HTML}{1B6FB8}   
\definecolor{tgtcol}{HTML}{D81E05}    
\title{GroundGUI: GUI Grounding by\\ High-Quality Data and\\ Distillation-to-Reinforcement Learning}

\author{%
Qingsong Xie\textsuperscript{1$*$}, Jiahao Tian\textsuperscript{1,2$*$}, Jiacheng Tang\textsuperscript{1}, Zhifeng Li\textsuperscript{1$\dagger$}, Madalina Ciobanu\textsuperscript{1}, 
Navan~Preet~Singh\textsuperscript{1}, Qingqing Mao\textsuperscript{1,3$\dagger$}, Ritankar Das\textsuperscript{1,3}\\
\textsuperscript{1}Prentis AI, San Francisco, CA \qquad \textsuperscript{2}Westlake University, Hangzhou, China\\
\rule{0pt}{10pt}\textsuperscript{3}Titan Holdings, San Francisco, CA
}

\paperfinalcopy 

\begin{document}

\maketitle
{\renewcommand{\thefootnote}{}\footnotetext{%
\textsuperscript{$*$}Equal contribution: \texttt{qingsongxie@sjtu.edu.cn}\\
\hspace*{1.8em}\textsuperscript{$\dagger$}Corresponding authors: 
\texttt{qmao@prentis.ai}, \texttt{zhifeng.li@titanresearchlabs.ai}.}}

\begin{abstract}
Constructing robust GUI agents relies on multimodal visual grounding—the ability to precisely map natural language intents to actionable on-screen components. Despite abundant desktop, web, and mobile benchmarks, high-fidelity resources tailored for diverse environments are surprisingly scarce. To bridge this gap, we introduce the Screen Referring Grounding Dataset (ScreenRef), a comprehensive GUI grounding dataset consisting of GUI screenshots paired with natural language referring expressions and box or point annotations. Spanning desktop, web, and mobile domains, ScreenRef comprises 586K screenshots and over 633K referring expression instances, covering tasks such as text recognition, visual matching, spatial understanding, and refusal scenarios. To boost the model’s grounding performance, we propose the Distillation-to-Reinforcement Learning (D2RL) method to address the limited generalization of supervised fine-tuning and the sparse reward problem in reinforcement learning. Utilizing ScreenRef, we train Qwen3-VL-8B-Instruct via sequential SFT and D2RL training, yielding GroundGUI-8B. GroundGUI-8B achieves 95.2\% on  ScreenSpot-V2, 63.9\% on ScreenSpot-Pro, 65.4\% on OSWorld-G, 86.8\% on MMBench-GUI L2, and 38.3\% on UI-Vision. Among 8B-scale models, it produces substantial gains over models trained on open-source datasets, with performance comparable or even superior to models trained on closed-source data. Code, models, and datasets are
available at https://github.com/PrentisAI/GroundGUI.
\end{abstract}

\section{Introduction}
Graphical User Interfaces (GUIs)~\citep{cheng2024seeclick} are ubiquitous in modern human-computer and mobile interactions. Visual GUI grounding (mapping language instructions to pixel coordinates of interactive elements) is a foundational capability. Robust cross-platform grounding is challenging due to interface complexity and contextual diversity. Modern applications often have high-resolution canvases with dense layouts and subtle visual disambiguations, complicating spatial localization. GUI agents degrade on domain-specific software due to limited grasp of specialized icon semantics. Data underpins model capability, but constructing high-quality grounding datasets with broad coverage is challenging~\citep{xu2024aguvis,xie2025scaling}. Mobile-Agent-v3.5~\citep{xu2026mobile}, UI-Venus-1.5~\citep{team2026ui}, UI-TARS-2~\citep{wang2025ui}, and MAI-UI~\citep{zhou2025maiui} perform well with proprietary large-scale data, unavailable to the community. GroundCUA~\citep{feizi2026grounding} and ScaleCUA~\citep{liu2026scalecua} release their data, but models trained on them are suboptimal due to limited diversity/accuracy.  

We introduce ScreenRef, a large-scale GUI grounding dataset (586K screenshots and 633K referring expressions), with text recognition, visual matching, spatial understanding, and refusal tasks. It covers desktop, web, and mobile platforms, with image resolutions ranging from 0.1 to 8.3M pixels. Multi-stage filtering/verification yield verified data for robust cross-platform grounding.

Supervised fine-tuning (SFT) is commonly used to train grounding models, but overfits to specific domains~\citep{luo2025gui}.
Reinforcement learning (RL) works~\citep{tang2026gui,li2026gui,liu2026infigui,lu2026ui} use Group Relative Policy Optimization (GRPO)~\citep{shao2024deepseekmath} to improve generalization, but suffer from sparse signals on hard samples where all rollouts
receive zero reward. Inspired by  on-policy self-distillation (OPSD)~\citep{zhao2026self} which provides dense token-level supervision, we propose \textbf{Distillation-to-Reinforcement Learning (D2RL)}, a two-stage post-training framework: OPSD followed by REINFORCE Leave-One-Out (RLOO)~\citep{ahmadian2024back}. OPSD tailored for GUI grounding delivers rich token-level supervision for precise coordinate generation from a teacher that sees the ground-truth bounding box, providing a stronger starting policy before RL. RLOO then enhances model generalization and localization accuracy. Through  the  proposed ScreenRef and D2RL method,  we obtain GroundGUI for precise GUI grounding
across desktop, web, and mobile. GroundGUI-8B achieves 95.2\% on ScreenSpot-V2~\citep{wu2025atlas}, 63.9\% on ScreenSpot-Pro~\citep{li2025screenspotpro}, 65.4\% on OSWorld-G~\citep{xie2025scaling}, 86.8\% on MMBench-GUI L2~\citep{wang2025mmbench}, and 38.3\% on UI-Vision, remarkably outperforming the models trained on open-source datasets. GroundGUI-8B also surpasses
Holo2-8B~\citep{hai2025holo2modelfamily} and Step-GUI-8B~\citep{yan2025step} by a large margin, both of which are trained on closed-source data.

\textbf{Core contributions.}  
(1) \textbf{ScreenRef}, a high-quality GUI grounding dataset with
586K screenshots and 633K referring expressions, providing text recognition,
visual matching, spatial understanding, and refusal supervision for robust GUI agents. 
(2) \textbf{D2RL}, which proposes OPSD before RL to shape a better initial policy, unleashing the potential of RL
for GUI grounding, enhancing model generalization and grounding accuracy. 
(3) \textbf{GroundGUI-8B}, trained on our proposed data and approach.
GroundGUI-8B achieves state-of-the-art or competitive performance among models with
8B parameters or fewer, surpassing several 32B-scale models, e.g.,
ScaleCUA-32B, OpenCUA-32B and GTA1-32B. We release the data, code, and model to the community to facilitate future research.

\section{Related Works}

\textbf{GUI Agent.}
Computer-use agents interact with interfaces, from simple commands
to multi-step tasks. CogAgent~\citep{hong2024cogagent},
ShowUI~\citep{lin2025showui}, Ferret-UI~\citep{you2024ferret},  OS-Genesis~\citep{sun2025genesis}, GUI-Actor~\citep{wu2026gui}, and UI-TARS~\citep{qin2025ui} improve interaction via SFT. OS-ATLAS~\citep{wu2025atlas} and UGround~\citep{gou2025uground} scale  data to map language to UI
components. These methods are data inefficient, particularly in complex
desktop and web environments. RL approaches, inspired by GRPO~\citep{shao2024deepseekmath} (GUI-R1~\citep{luo2025gui}, GUI-G$^2$~\citep{tang2026gui}, GUI-C$^2$~\citep{li2026gui}, InfiGUI-G1~\citep{liu2026infigui}), improve grounding. Still, reliably grounding instructions to hard  on-screen elements remains a bottleneck due to sparse signals on difficult samples where all rollouts
generate erroneous responses.  Motivated by OPSD, an approach that provides online token‑level supervision via a single rollout, we introduce D2RL, a two‑stage training framework. In the preparatory phase, the framework performs OPSD tailored for GUI grounding to bootstrap subsequent reinforcement‑learning optimization.

\textbf{GUI Grounding Datasets.} GUI grounding datasets span mobile, web, and desktop environments. Mobile datasets (e.g., RICO~\citep{deka2017rico}, UIBert~\citep{bai2021uibert}, AITW~\citep{rawles2023androidinthewild}, AITZ~\citep{zhang2024android}, AMEX~\citep{chai2025amex}) provide element-level supervision in standardized layouts, which simplifies extraction but offers limited layout density and visual variation vs. desktop software. Web-focused datasets (e.g., SeeClick~\citep{cheng2024seeclick}, WebUI~\citep{wu2023webui}, UGround, and AGUVIS-G~\citep{xu2024aguvis}) scale collection by harvesting HTML/DOM structures. These favor text-bearing elements over icon-only controls.
High-quality desktop datasets remain scarce and hard to curate. OS-ATLAS~\citep{wu2025atlas} extracts desktop elements from accessibility trees; inconsistent accessibility APIs introduce label noise. JEDI~\citep{xie2025scaling} scales dataset size via synthetic interface generation, but its simplified screens fail to capture desktop complexity. Recent efforts such as GroundCUA~\citep{feizi2026grounding}, ScaleCUA~\citep{liu2026scalecua}, and Wild2GUI~\citep{xiong2026video2gui} offer dense annotations, cross-platform coverage, or video-synthesized data, but annotation inaccuracies or sparse diversity constrain performance. To overcome the drawbacks of the existing datasets, we propose a high-quality ScreenRef dataset with high diversity  via multi-stage filtering and verification, covering web, mobile, and desktop platforms.

\makeatletter
\@ifundefined{textcolor}
  {\providecommand{\SR@tint}[2]{#2}}
  {\definecolor{SRgreen}{RGB}{0,140,60}%
   \definecolor{SRred}{RGB}{160,20,20}%
   \definecolor{SRblue}{RGB}{20,80,200}%
   \providecommand{\SR@tint}[2]{\textcolor{#1}{#2}}}

\@ifundefined{ding}
  {\@ifundefined{checkmark}
     {\providecommand{\SR@ck}{\ensuremath{\surd}}}
     {\providecommand{\SR@ck}{\ensuremath{\checkmark}}}%
   \providecommand{\SR@cr}{\ensuremath{\times}}}
  {\providecommand{\SR@ck}{\ding{51}}\providecommand{\SR@cr}{\ding{55}}}
\@ifundefined{textcolor}
  {\providecommand{\SRyes}{\SR@ck}\providecommand{\SRno}{\SR@cr}%
   \providecommand{\SRcit}[1]{#1}}
  {\providecommand{\SRyes}{\SR@tint{SRgreen}{\SR@ck}}%
   \providecommand{\SRno}{\SR@tint{SRred}{\SR@cr}}%
   \providecommand{\SRcit}[1]{\SR@tint{SRblue}{#1}}}
\@ifundefined{resizebox}
  {\providecommand{\SRfit}[1]{#1}}
  {\providecommand{\SRfit}[1]{\resizebox{\linewidth}{!}{#1}}}
\makeatother

\begin{table}[t]
\centering
\caption{Comparison with existing GUI grounding datasets.
Instruction counts follow each dataset's original reporting unit and are not directly comparable
across datasets; details in Appendix~\ref{app:dataset-comparison}.}
\label{tab:grounding-datasets}

\small
\setlength{\tabcolsep}{5pt}
\renewcommand{\arraystretch}{0.90}
\SRfit{%
\begin{tabular}{l ccc cc c c}
\toprule
 & \multicolumn{3}{c}{\textbf{Platform}} & \multicolumn{3}{c}{\textbf{Scale \& Statistics}} & \\
\cmidrule(lr){2-4}\cmidrule(lr){5-7}
\textbf{Dataset} & \textbf{Web} & \textbf{Mobile} & \textbf{Desktop}
 & \textbf{Images} & \textbf{Instr.} & \textbf{Res Range} & \textbf{Refusal} \\
\midrule
OS-ATLAS~\citep{wu2025atlas}                  & \SRyes & \SRyes & \SRyes & 1.85M & ---   & (0.5, 5.2) & \SRno  \\
FineWeb~\citep{penedo2024fineweb}             & \SRyes & \SRno & \SRno & 1.4M  & ---   & (2.1, 2.1) & \SRno  \\
UGround~\citep{gou2025uground}                & \SRyes & \SRno & \SRno & 773K  & ---   & (0.4, 1.9) & \SRno  \\
Aria-UI~\citep{yang2025ariaui}                & \SRyes & \SRyes & \SRyes & 295K  & ---   & (1.3, 1.9) & \SRno  \\
SeeClick~\citep{cheng2024seeclick}            & \SRyes & \SRno & \SRno & 270K  & ---   & (2.1, 2.1) & \SRno  \\
\addlinespace[3pt]
GroundCUA~\citep{feizi2026grounding}          & \SRno & \SRno & \SRyes & 55K   & 700K  & (0.4, 7.0) & \SRno  \\
AMEX~\citep{chai2025amex}                     & \SRno & \SRyes & \SRno & 101K  & ---   & (0.9, 4.5) & \SRno  \\
\addlinespace[3pt]
JEDI~\citep{xie2025scaling}                   & \SRyes & \SRyes & \SRyes & 575K  & 4M    & (0.9, 2.1) & \SRyes  \\
MolmoPoint-GUISyn~\citep{clark2026molmopoint} & \SRyes & \SRyes & \SRyes & 36K   & 10M   & (0.9, 8.3) & \SRno  \\
\addlinespace[3pt]
AGUVIS-G~\citep{xu2024aguvis}                 & \SRyes & \SRyes & \SRyes & 452K  & 1.04M & (0.5, 2.1) & \SRno  \\
UIBert~\citep{bai2021uibert}                  & \SRno & \SRyes & \SRno & 57K   & ---   & (0.5, 2.1) & \SRno  \\
RICOSCA~\citep{li2020ricosca}                 & \SRno & \SRyes & \SRno & 18K   & ---   & (0.5, 2.1) & \SRno  \\
Widget Caption~\citep{li2020widget}           & \SRno & \SRyes & \SRno & 14K   & ---   & (0.5, 2.1) & \SRno  \\
\midrule
\textsc{ScreenRef} (ours)                     & \SRyes & \SRyes & \SRyes & 586K  & 633K  & \textbf{(0.1, 8.3)} & \SRyes \\
\bottomrule
\end{tabular}}
\end{table}

\begin{figure}[!t]
\centering
\includegraphics[width=\textwidth]{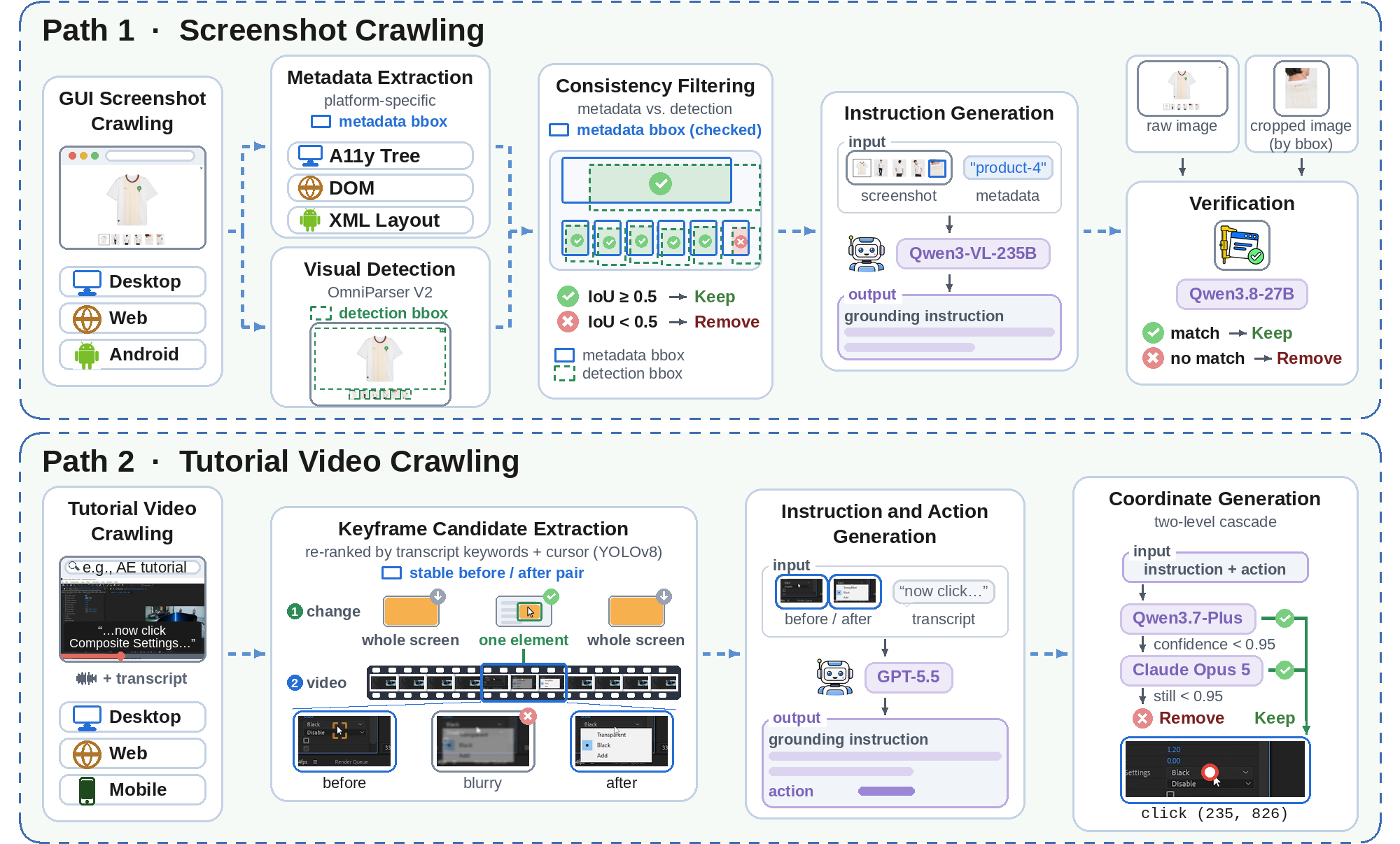}
\caption{\textbf{Overview of the ScreenRef data construction pipeline.}
\textbf{Path 1} consists of screenshot crawling, metadata filtering, instruction generation, and verification stages.
\textbf{Path 2} contains tutorial video crawling, keyframe candidate extraction, joint instruction and action generation, and coordinate generation stages.}
\label{fig:screenref-pipeline}

\end{figure}

\section{ScreenRef Dataset}
\label{sec:screenref}

\subsection{Screenshot Crawling}
We use platform-specific approaches for GUI metadata extraction (Figure~\ref{fig:screenref-pipeline}, Path 1). For desktop, Accessibility Trees (A11y Trees) provide interface layouts and UI element coordinates. Document Object Model (DOM) structures offer element information for web browsers. Parsed XML layout files provide UI
element locations for Android applications. Metadata contains  bounding boxes,
categories, and text. 
However, some metadata contains noisy bounding box attributes. For these cases, we adopt OmniParser V2~\citep{yu2026omniparser} to reduce noise by comparing the metadata bounding boxes with the detection results of OmniParser. The bounding boxes with an IoU lower than 0.5 are removed. Figure \ref{fig:element-metadata} in Appendix~\ref{app:element-metadata} shows the annotation for the screenshot.
 Qwen3-VL-235B-A22B-Instruct~\citep{bai2025qwen3} is then employed to produce grounding instructions with annotated bounding boxes, categories, and surrounding context. To further verify the correctness of the generated grounding pairs, we adopt Qwen3.8-27B~\citep{qwen38} to judge whether the answer matches the instruction given the raw image and the bounding box crop.
 Through this pipeline, we generate three primary types of instructions, including text recognition, visual matching, and spatial understanding.  Text recognition denotes grounding a target element uniquely designated by readable on-screen text. Visual matching refers to identifying the target element by its visual attributes (icon noun, shape, or color). Spatial understanding represents identifying a target element purely through its spatial relationship, whether relative to an anchor element or bounded within a named container. Qualitative examples are provided in
Appendix~\ref{app:taxonomy-examples}.

\subsection{Tutorial Video Crawling}
To collect real-world interfaces, we complement
screenshots  with interaction frames from public tutorial videos. To this end,
search queries are automatically generated to crawl video metadata on YouTube.
Qwen3.6-27B~\citep{qwen3.6-27b} is then used to  filter out non-tutorial video metadata, followed by video downloading based on the preserved metadata.
For subtitle-free videos, FFmpeg is exploited to extract audio for transcript generation. 
To produce grounding samples, we design a 
pipeline  converting tutorial videos into trainable $(\text{instruction},\ \text{observation},\ \text{action},\ \text{coordinate})$ tuples via 3 sequential stages: (i) \textbf{Keyframe Candidate Extraction.} It  computes block-level frame differences to pinpoint timestamps of local screen updates. To prevent global screen transitions (shot cuts, page scrolling, video playback) from dominating, we down-weight full-frame edits, letting fine-grained interactions compete. Candidates are re-ranked using action verbs extracted from the transcript alongside cursor coordinates detected via YOLOv8~\citep{reis2023real} as cheap priors, yielding at most 60 filtered candidates/video. Instead of capturing the immediate moment of change (often with motion blur or mid-animation states), we perform a forward-and-backward search around each timestamp to identify the nearest static frames, constructing a clean pre-/post-change frame pair. (ii) \textbf{Joint Instruction and Action Annotation.} Each candidate frame pair, with absolute timestamps, is concatenated with its transcript segment and video metadata into a single GPT-5.5 API prompt. This pass performs both task segmentation and action annotation, outputting a sequence of temporal actions. Each action includes its interaction type, parameter values, and an element-level referring expression grounded strictly in visible screen evidence (referencing UI controls solely by their visible text labels, icon shapes, or explicit functional descriptions).  (iii) \textbf{Post-Hoc Visual Grounding.} Taking the pre-action frame as the visual observation, the target coordinate is resolved via action-type-conditioned routing. A single call to Qwen3.7-Plus~\citep{qwen37plus} predicts target presence, a bounding box/coordinate, and a confidence score. If confidence falls below a threshold $\tau=0.95$, the proposal is escalated to Claude Opus 5 for re-grounding. Samples with secondary confidence below $\tau$ are discarded. To fill the critical gap of handling infeasible grounding queries in existing datasets, we construct refusal samples via strategically random-pairing instructions and frames. We then employ the Qwen3.8-27B model to identify and verify genuinely infeasible grounding instances.

\subsection{Dataset Statistics}
Through dual-mode pipelines, we obtain ScreenRef, comprising 585,981 screenshots and 633,369 instructions (Instr), with  284,531 grounding samples from pathway 1 and the remaining from pathway 2.  As listed in Table \ref{tab:grounding-datasets}, ScreenRef features multiple platforms and  a wider resolution range (Res Range) than the existing datasets. ScreenRef and JEDI~\citep{xie2025scaling} are the only two datasets that include refusal samples.
 The distribution of the 4 task types across desktop, web, and mobile environments is given in Table~\ref{tab:data-capability}.

\begin{figure}[h]
\centering
\includegraphics[width=\textwidth]{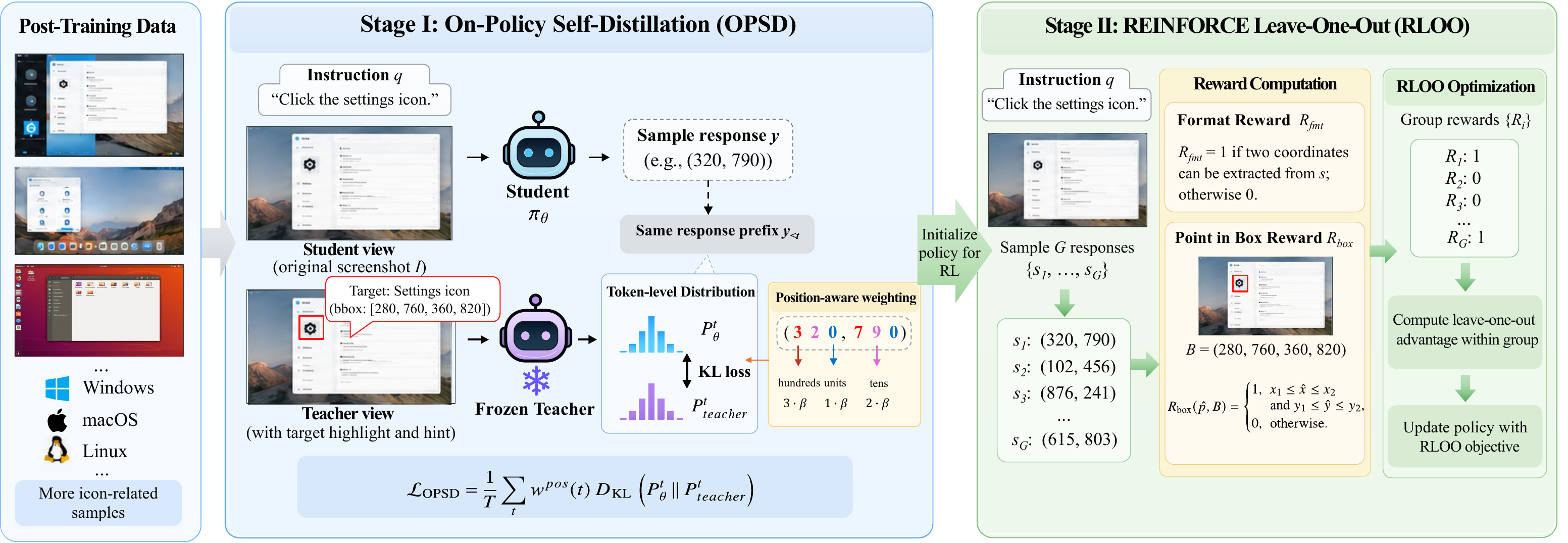}
\caption{\textbf{Overview of the D2RL method.} The format check is a gate, not a reward term.}
\label{fig:rl}
\end{figure}

\section{Distillation-to-Reinforcement Learning}
\label{sec:opsd_rloo}
SFT first fits the model to ScreenRef, improving grounding across diverse environments. 
RL then directly optimizes the task-level click reward. However, RL strongly depends on its starting policy. On challenging samples such as small icons or spatial-relation queries, all sampled rollouts may fail and receive zero reward, providing little learning signal. We thus propose \textbf{Distillation-to-Reinforcement Learning (D2RL)}, which applies on-policy self-distillation (OPSD)~\citep{zhao2026self} as an explicit preparation stage: dense teacher supervision on the student's own trajectories shapes a better starting policy before RL optimizes grounding success under the original screenshot (Table \ref{tab:rl-comparison} confirms the benefit of the stronger initial policy provided by OPSD).

\noindent\textbf{Stage I: On-Policy Self-Distillation.}
SFT learns from labeled responses, while RL samples from the model's own policy. OPSD bridges the two by matching the teacher's token distribution on student-generated prefixes via reverse KL divergence, providing guidance on the responses the model currently explores. Specifically, given a screenshot $I$ and an instruction $q$, the student policy $\pi_\theta$ samples a coordinate response $s$ from the original screenshot. Inspired by  GUI-SD~\citep{zhang2026learn}, the teacher is a frozen copy of the SFT policy; during OPSD only, it receives the screenshot with the ground-truth target box highlighted together with a textual localization hint (the target name and its box), whereas the student receives only the original screenshot and the instruction. The teacher is not used at inference. Both models score the same student-sampled prefix, so the teacher provides dense token-level guidance along the student’s rollout. Since errors in higher-order coordinate digits induce much larger spatial deviations than errors in
lower-order digits, we apply position-aware weighting during distillation: we set 
$w^{\mathrm{pos}}(t)=\beta \cdot k_t$ for coordinate-digit tokens and
$w^{\mathrm{pos}}(t)=1$ otherwise; $k_t$ denotes the significance of the digit (e.g., $1$ for units,
$2$ for tens, and $3$ for hundreds), and $\beta$ controls the weighting
strength.
The OPSD objective is then
\begin{equation}
\mathcal L_{\mathrm{OPSD}}
=
\frac{1}{T}\sum_t
w^{\mathrm{pos}}(t)
D_{\mathrm{KL}}
\left(
P_{\theta}^{t}
\,\|\, 
P_{\mathrm{teacher}}^{t}
\right),
\end{equation}
where $P_{\theta}^{t}$ and $P_{\mathrm{teacher}}^{t}$ denote the student and
teacher token distributions at decoding step $t$. Stage I aims to improve the RLOO starting policy, not only pre-RL accuracy. Dense privileged supervision shifts the student's trajectories toward target-consistent coordinates, reducing sparse reward reliance. Position-aware weighting emphasizes coordinate decisions that matter for localization.

\noindent\textbf{Stage II: REINFORCE Leave-One-Out.}
Starting from the OPSD-trained policy, we further optimize the grounding policy with the RLOO method~\citep{ahmadian2024back} under the original screenshot observations. Given a screenshot $I$ and an instruction $q$, the policy $\pi_\theta$ generates a coordinate response $s$. Each response is evaluated using a simple point-in-box reward. 

If the response cannot be parsed into a click point, it receives a reward of 0; the format check in Figure~\ref{fig:rl} thus acts as a gate rather than an additive reward term. Otherwise, the parsed point $\hat{\mathbf p}=(\hat{x},\hat{y})$ is evaluated in the normalized coordinate space. Let $B=(x_1,y_1,x_2,y_2)$ denote the target box, where $(x_1,y_1)$ and $(x_2,y_2)$ are its top left and bottom right corners. The point-in-box reward is
\begin{equation}
R_{\mathrm{box}}(\hat{\mathbf p},B)=
\begin{cases}
1, & x_1\leq \hat{x}\leq x_2 \text{ and } y_1\leq \hat{y}\leq y_2,\\
0, & \text{otherwise}.
\end{cases}
\label{eq:box_reward}
\end{equation}

For each screenshot and instruction pair, we sample $G$ responses $\{s_i\}_{i=1}^{G}$. Following RLOO, the reward of each rollout is compared with the average reward of the other rollouts in the same group, providing a leave-one-out baseline without a separate critic. Let $R_i$ denote the (point-in-box) reward of $s_i$. The policy gradient is given by
\begin{equation}
\nabla_\theta J(\pi_\theta)
=
\frac{1}{G}\sum_{i=1}^{G}
\left(
R_i
-
\frac{1}{G-1}\sum_{j\neq i} R_j
\right)
\nabla_\theta \log \pi_\theta(s_i \mid I,q).
\label{eq:rloo}
\end{equation}

The 2 stages provide complementary feedback. OPSD uses privileged information to guide student token distribution, including failed rollouts with identical click rewards, and prepares a better starting policy for RLOO, which uses task-level success to reweight responses under the original screenshot.

\section{Experimental Results}
\subsection{Implementation Details}
\label{sec:impl}
SFT is performed on 4 B300 GPUs, using a global batch size of 32 and a learning rate of $1\times10^{-6}$. The model is trained for 2 epochs using the AdamW~\citep{loshchilov2018decoupled} optimizer. In the second epoch, the loss weights for text recognition and refusal samples are reduced to 0.2.
Starting from the SFT model, we post-train with D2RL on 12K grounding samples derived from GroundCUA, excluding screenshots identical to those in the UI-Vision benchmark. Training runs for 2 epochs (one for OPSD and one for RLOO) on 8 B300 GPUs. Both stages use AdamW with a learning rate of $1\times10^{-6}$, a cosine schedule, and a batch size of 42, without a KL penalty. OPSD matches a frozen teacher's token distributions on the policy's own rollouts, followed by RLOO with 8 rollouts per prompt and a binary point-in-box reward.

\subsection{Evaluation Benchmarks}
\label{sec:bench}

\begin{table*}[!t]
\footnotesize
\centering
\caption{Performance on \textbf{ScreenSpot-V2}.
\textbf{Bold} values = best results.}
\label{tab:screenspot_v2}
\renewcommand{\arraystretch}{0.90}
\setlength{\tabcolsep}{0pt}

\begin{tabular*}{\linewidth}{@{\extracolsep{\fill}}l cc cc cc c@{}}
\toprule
\multirow{2}{*}{\textbf{Model}}
& \multicolumn{2}{c}{\textbf{Mobile}}
& \multicolumn{2}{c}{\textbf{Desktop}}
& \multicolumn{2}{c}{\textbf{Web}}
& \multirow{2}{*}{\textbf{Avg.}} \\
\cmidrule(lr){2-3}
\cmidrule(lr){4-5}
\cmidrule(lr){6-7}
& Text & Icon. & Text & Icon. & Text & Icon. & \\
\midrule

\rowcolor{secrow}
\multicolumn{8}{l}{\textit{Open-Source Models}} \\

Qwen3-VL-2B~\citep{bai2025qwen3}
& 95.5 & 82.0 & 95.4 & 73.6 & 89.7 & 76.4 & 86.7 \\

Phi-Ground-7B~\citep{phiground2024}
& 90.2 & 76.4 & 93.6 & 75.9 & 96.5 & 62.0 & 83.8 \\

OS-ATLAS-7B~\citep{wu2025atlas}
& 95.2 & 75.8 & 90.7 & 63.6 & 90.6 & 77.3 & 85.1 \\

UGround-V1-7B~\citep{gou2025uground}
& 83.6 & 90.5 & 85.8 & 86.3 & 95.5 & 83.2 & 87.7 \\

UI-TARS-1.5-7B~\citep{qin2025ui}
& 92.2 & 81.5 & 91.0 & 84.2 & 95.5 & 84.5 & 89.0 \\

SE-GUI-7B~\citep{yuan2025segui}
& 99.3 & 89.1 & 96.4 & 78.6 & 92.7 & 81.3 & 90.8 \\

UI-TARS-7B~\citep{qin2025ui}
& 96.9 & 89.1 & 95.4 & 85.0 & 93.6 & 85.2 & 91.6 \\

Qwen3-VL-8B~\citep{bai2025qwen3}
& 97.9 & 84.8 & 95.9 & 87.9 & 95.7 & 83.7 & 91.7 \\

GUI-Actor-7B~\citep{wu2026gui}
& 97.6 & 88.2 & 96.9 & 85.7 & 93.2 & 86.7 & 92.1 \\

OpenCUA-7B~\citep{wang2026opencua}
& -- & -- & -- & -- & -- & -- & 92.3 \\

GTA1-7B~\citep{gta1_2025}
& 99.0 & 88.6 & 94.9 & 89.3 & 92.3 & 86.7 & 92.4 \\

GUI-Owl-7B~\citep{mobileagentv3_2025}
& 99.0 & 92.4 & 96.9 & 85.0 & 93.6 & 85.2 & 92.8 \\

GUI-G$^2$-7B~\citep{tang2026gui}
& 98.3 & 91.9 & 95.4 & 89.3 & 94.0 & 87.7 & 93.3 \\

InfiGUI-G1-7B~\citep{liu2026infigui}
& 99.0 & 91.9 & 94.3 & 82.1 & \textbf{97.9} & 89.2 & 93.5 \\

UI-Venus-7B~\citep{uivenus2025}
& 99.0 & 90.0 & 96.9 & 90.7 & 96.2 & 88.7 & 94.1 \\

MAI-UI-8B~\citep{zhou2025maiui}
& 99.3 & 89.1 & 99.0 & 92.1 & \textbf{97.9} & 91.1 & 95.2 \\

GUI-Owl-1.5-8B~\citep{xu2026mobile}
& 97.4 & 90.5 & 96.4 & 90.7 & 94.2 & 89.7 & 93.7 \\

Holo2-8B~\citep{hai2025holo2modelfamily}
& -- & -- & -- & -- & -- & -- & 93.2 \\

MolmoPoint-8B~\citep{clark2026molmopoint}
& -- & -- & -- & -- & -- & -- & 93.4 \\

Step-GUI-8B~\citep{yan2025step}
& -- & -- & -- & -- & -- & -- & 95.1 \\

COPC-8B~\citep{wang2026measure}
& -- & -- & -- & -- & -- & -- & 91.3 \\

\midrule

Qwen3-VL-32B~\citep{bai2025qwen3}
& 96.2 & 90.0 & 97.4 & 85.0 & 95.7 & 89.7 & 93.0 \\

GUI-Owl-32B~\citep{mobileagentv3_2025}
& 98.6 & 90.0 & 97.9 & 87.8 & 94.4 & 86.7 & 93.2 \\

OpenCUA-32B~\citep{wang2026opencua}
& -- & -- & -- & -- & -- & -- & 93.4 \\

ScaleCUA-32B~\citep{liu2026scalecua}
& 98.6 & 91.9 & 99.0 & 90.0 & 94.4 & 91.6 & 94.7 \\

GTA1-32B~\citep{gta1_2025}
& \textbf{99.7} & 90.5 & 99.0 & \textbf{94.3}
& 95.7 & 90.1 & 95.2 \\

UI-Venus-72B~\citep{uivenus2025}
& \textbf{99.7} & \textbf{93.8} & 95.9 & 90.0
& 96.2 & \textbf{92.6} & \textbf{95.3} \\

UI-TARS-72B~\citep{qin2025ui}
& 94.8 & 86.3 & 91.2 & 87.9 & 91.5 & 87.7 & 90.3 \\

\midrule

GroundGUI-8B
& 98.6 & 90.0 & \textbf{99.5} & 92.1
& 96.2 & \textbf{92.6} & 95.2 \\

\bottomrule
\end{tabular*}
\end{table*}

\begin{table*}[!t]
\footnotesize
\centering
\caption{Performance comparison on \textbf{MMBench-GUI L2} benchmark.
\textbf{Bold} values = best results.}
\label{tab:mmbench-gui-l2}
\setlength{\tabcolsep}{1.95pt}
\renewcommand{\arraystretch}{0.90}

\begin{tabular}{l cc cc cc cc cc cc c}
\toprule
\multirow{2}{*}{\textbf{Model}}
& \multicolumn{2}{c}{\textbf{Windows}}
& \multicolumn{2}{c}{\textbf{macOS}}
& \multicolumn{2}{c}{\textbf{Linux}}
& \multicolumn{2}{c}{\textbf{iOS}}
& \multicolumn{2}{c}{\textbf{Android}}
& \multicolumn{2}{c}{\textbf{Web}}
& \multirow{2}{*}{\textbf{Avg.}} \\
\cmidrule(lr){2-3}
\cmidrule(lr){4-5}
\cmidrule(lr){6-7}
\cmidrule(lr){8-9}
\cmidrule(lr){10-11}
\cmidrule(lr){12-13}
& Bas. & Adv.
& Bas. & Adv.
& Bas. & Adv.
& Bas. & Adv.
& Bas. & Adv.
& Bas. & Adv.
& \\
\midrule

Qwen3-VL-2B~\citep{bai2025qwen3}
& 81.9 & 0.0
& 80.3 & 46.2
& 67.5 & 0.0
& 90.8 & 0.0
& 91.0 & 0.0
& 88.1 & 0.0
& 46.5 \\

OS-ATLAS-7B~\citep{wu2025atlas}
& 36.9 & 18.8
& 44.4 & 21.7
& 31.4 & 13.3
& 74.8 & 48.8
& 69.6 & 46.8
& 61.3 & 35.4
& 41.4 \\

Aguvis-7B~\citep{xu2024aguvis}
& 37.3 & 21.7
& 48.1 & 33.3
& 33.5 & 25.0
& 67.5 & 65.2
& 61.0 & 51.0
& 61.6 & 45.5
& 45.7 \\

UI-TARS-1.5-7B~\citep{qin2025ui}
& 68.3 & 39.0
& 69.0 & 44.5
& 64.4 & 37.8
& 88.5 & 69.4
& 90.5 & 69.3
& 81.0 & 56.5
& 64.3 \\

UGround-V1-7B~\citep{gou2025uground}
& 66.8 & 39.0
& 71.3 & 48.6
& 56.5 & 31.1
& 92.7 & 70.9
& 93.5 & 71.0
& 88.7 & 64.6
& 65.7 \\

GUI-Actor-7B~\citep{wu2026gui}
& 80.8 & 55.1
& 81.4 & 60.4
& 64.9 & 41.8
& 94.3 & 82.7
& 93.5 & 79.7
& 89.7 & 72.1
& 76.5 \\

SE-GUI-7B~\citep{yuan2025segui}
& 77.5 & 57.7
& 77.1 & 60.7
& 68.6 & 44.9
& 95.5 & 80.0
& 95.5 & 83.7
& 89.7 & 68.8
& 76.6 \\

Qwen3-VL-8B~\citep{bai2025qwen3}
& 88.6 & 61.8
& 85.5 & 69.1
& 74.9 & 53.1
& 95.2 & 82.4
& 95.5 & 84.5
& \textbf{96.8} & 72.1
& 81.3 \\

GTA1-7B~\citep{gta1_2025}
& 76.8 & 57.4
& 80.3 & 63.9
& 68.6 & 53.6
& 93.9 & 83.3
& 96.3 & 84.5
& 90.3 & 74.7
& 78.5 \\

GUI-G$^2$-7B~\citep{tang2026gui}
& 79.7 & 55.1
& 79.7 & 64.7
& 69.6 & 50.0
& 95.2 & 82.7
& 96.6 & 85.4
& 91.9 & 75.6
& 78.8 \\

GUI-Owl-7B~\citep{mobileagentv3_2025}
& 86.4 & 61.8
& 81.7 & 64.5
& 74.4 & 61.7
& 94.9 & 83.0
& 95.8 & 83.7
& 93.2 & 72.7
& 80.5 \\

InfiGUI-G1-7B~\citep{liu2026infigui}
& 82.7 & 61.8
& 83.8 & 63.9
& 72.3 & 52.0
& 94.9 & 89.4
& 95.2 & 85.6
& 93.5 & 76.3
& 80.8 \\

Step-GUI-8B~\citep{yan2025step}
& - & - & - & - & - & - & - & - & - & - & - & - & 85.6 \\

COPC-8B~\citep{wang2026measure}
& - & - & - & - & - & - & - & - & - & - & - & - & 78.4 \\

\midrule

GUI-Owl-32B~\citep{mobileagentv3_2025}
& 85.6 & 65.1
& 84.9 & 67.1
& 77.0 & \textbf{63.3}
& 95.2 & 85.5
& 96.1 & 87.0
& 95.5 & 80.8
& 83.0 \\

GTA1-32B~\citep{gta1_2025}
& 82.3 & 66.9
& 89.0 & 74.0
& 73.3 & 52.0
& 96.2 & 88.2
& 95.8 & 88.5
& 95.2 & 79.9
& 83.4 \\

Qwen3-VL-32B~\citep{bai2025qwen3}
& \textbf{93.4} & \textbf{71.3}
& \textbf{92.8} & 74.3
& 78.0 & 56.1
& 95.5 & 88.8
& 97.2 & 88.5
& 92.6 & 78.6
& 85.3 \\
\midrule
GroundGUI-8B (Ours)
& 92.3 & 69.9
& 89.0 & \textbf{80.3}
& \textbf{81.7} & 61.7
& \textbf{96.8} & \textbf{89.7}
& \textbf{97.5} & \textbf{91.3}
& 94.8 & \textbf{82.5}
& \textbf{86.8} \\

\bottomrule
\end{tabular}
\end{table*}

\begin{table*}[!t]
\footnotesize
\centering
\caption{Performance comparison on \textbf{ScreenSpot-Pro} benchmark.
\textbf{Bold} values = best results.}
\label{tab:screenspot-pro}
\setlength{\tabcolsep}{2pt}
\renewcommand{\arraystretch}{0.90}

\begin{tabular}{l cc cc cc cc cc cc c}
\toprule
\multirow{2}{*}{\textbf{Model}}
& \multicolumn{2}{c}{\textbf{CAD}}
& \multicolumn{2}{c}{\textbf{Dev.}}
& \multicolumn{2}{c}{\textbf{Creative}}
& \multicolumn{2}{c}{\textbf{Scientific}}
& \multicolumn{2}{c}{\textbf{Office}}
& \multicolumn{2}{c}{\textbf{OS}}
& \multirow{2}{*}{\textbf{Avg.}} \\
\cmidrule(lr){2-3}
\cmidrule(lr){4-5}
\cmidrule(lr){6-7}
\cmidrule(lr){8-9}
\cmidrule(lr){10-11}
\cmidrule(lr){12-13}
& Text & Icon
& Text & Icon
& Text & Icon
& Text & Icon
& Text & Icon
& Text & Icon
& \\
\midrule

Qwen3-VL-2B~\citep{bai2025qwen3}
& 31.0 & 15.6
& 55.2 & 11.7
& 59.1 & 16.1
& 64.6 & 22.7
& 72.3 & 34.0
& 59.8 & 23.6
& 41.9 \\

InfiGUI-3B~\citep{liu2026infigui}
& 50.8 & 25.0
& 64.9 & 20.0
& 51.5 & 16.8
& 68.8 & 32.7
& 70.6 & 32.1
& 49.5 & 19.7
& 45.2 \\

Ferret-UI Lite~\citep{you2024ferret}
& - & -
& - & -
& - & -
& - & -
& - & -
& - & -
& 53.3 \\

UI-TARS-7B~\citep{qin2025ui}
& 20.8 & 9.4
& 58.4 & 12.4
& 50.0 & 9.1
& 63.9 & 31.8
& 63.3 & 20.8
& 30.8 & 16.9
& 35.7 \\

Phi-Ground~\citep{phiground2024}
& 26.9 & 17.2
& 70.8 & 16.7
& 56.6 & 13.3
& 58.0 & 29.1
& 76.4 & 44.0
& 55.1 & 25.8
& 43.2 \\

GUI-Actor-7B~\citep{wu2026gui}
& 47.7 & 9.4
& 59.1 & 15.9
& 59.6 & 16.1
& 70.1 & 25.5
& 69.5 & 41.5
& 55.1 & 19.1
& 44.6 \\

SE-GUI-7B~\citep{yuan2025segui}
& 51.3 & 14.1
& 68.2 & 19.3
& 57.6 & 9.1
& 75.0 & 28.2
& 78.5 & 43.4
& 49.5 & 25.8
& 47.2 \\

GUI-G$^2$-7B~\citep{tang2026gui}
& 55.8 & 12.5
& 68.8 & 17.2
& 57.1 & 15.4
& 77.1 & 24.5
& 74.0 & 32.7
& 57.9 & 21.3
& 47.5 \\

Qwen3-VL-8B~\citep{bai2025qwen3}
& 46.7 & 10.9
& 79.2 & 23.4
& 68.2 & 14.0
& 73.6 & 30.0
& 76.3 & 30.2
& 65.4 & 21.3
& 49.9 \\

OpenCUA-7B~\citep{wang2026opencua}
& - & -
& - & -
& - & -
& - & -
& - & -
& - & -
& 50.0 \\

GTA1-7B~\citep{gta1_2025}
& 53.3 & 17.2
& 66.9 & 20.7
& 62.6 & 18.9
& 76.4 & 31.8
& 82.5 & \textbf{50.9}
& 48.6 & 25.9
& 50.1 \\

UI-Venus-7B~\citep{uivenus2025}
& 60.4 & 21.9
& 74.7 & 24.1
& 63.1 & 14.7
& 76.4 & 31.8
& 75.7 & 41.5
& 49.5 & 22.5
& 50.8 \\

InfiGUI-G1-7B~\citep{liu2026infigui}
& 57.4 & 23.4
& 74.7 & 24.1
& 64.6 & 18.2
& 80.6 & 31.8
& 75.7 & 39.6
& 57.0 & 29.2
& 51.9 \\

Holo2-8B~\citep{hai2025holo2modelfamily}
& - & -
& - & -
& - & -
& - & -
& - & -
& - & -
& 58.9 \\

MolmoPoint-8B~\citep{clark2026molmopoint}
& - & -
& - & -
& - & -
& - & -
& - & -
& - & -
& 60.2 \\

Step-GUI-8B~\citep{yan2025step}
& - & -
& - & -
& - & -
& - & -
& - & -
& - & -
& 62.6 \\

COPC-8B~\citep{wang2026measure}
& - & -
& - & -
& - & -
& - & -
& - & -
& - & -
& 58.7 \\

\midrule

Qwen3-VL-32B~\citep{bai2025qwen3}
& 60.4 & 28.1
& 69.5 & 22.1
& \textbf{75.8} & 25.2
& \textbf{84.7} & 25.5
& 85.9 & 43.4
& 62.6 & 15.7
& 54.9 \\

OpenCUA-32B~\citep{wang2026opencua}
& - & -
& - & -
& - & -
& - & -
& - & -
& - & -
& 55.3 \\

GUI-Owl-32B~\citep{mobileagentv3_2025}
& 62.4 & 28.1
& \textbf{84.4} & 39.3
& 65.2 & 18.2
& 82.6 & 39.1
& 81.4 & 39.6
& 70.1 & 36.0
& 58.0 \\

ScaleCUA-32B~\citep{liu2026scalecua}
& 60.4 & 39.1
& 75.3 & 35.2
& 73.2 & 30.8
& 76.4 & \textbf{46.4}
& 81.4 & 49.1
& 63.6 & 41.6
& 59.2 \\

UGround-V1-72B~\citep{gou2025uground}
& 16.8 & 4.7
& 55.8 & 4.8
& 54.0 & 10.5
& 70.8 & 22.7
& 61.0 & 18.9
& 40.2 & 7.9
& 34.5 \\

UI-TARS-72B~\citep{qin2025ui}
& 18.8 & 12.5
& 63.0 & 17.2
& 57.0 & 15.4
& 64.6 & 20.9
& 63.3 & 26.4
& 42.1 & 15.7
& 38.1 \\

UI-Venus-72B~\citep{uivenus2025}
& - & -
& - & -
& - & -
& - & -
& - & -
& - & -
& 61.9 \\

\midrule

GroundGUI-8B
& \textbf{67.5} & \textbf{40.6}
& \textbf{84.4} & \textbf{42.1}
& 74.2 & \textbf{35.0}
& 80.6 & 41.8
& \textbf{89.8} & \textbf{50.9}
& \textbf{71.0} & \textbf{44.9}
& \textbf{63.9} \\

\bottomrule
\end{tabular}
\end{table*}

\begin{table*}[!t]
\footnotesize
\centering
\caption{Performance comparison on \textbf{OSWorld-G} benchmark.
\textbf{Bold} values = best results.}
\label{tab:osworld-g}
\setlength{\tabcolsep}{2pt}
\renewcommand{\arraystretch}{0.90}

\begin{tabular}{l c c c c c c}
\toprule
\textbf{Model}
& \shortstack{\textbf{Text}\\\textbf{Matching}}
& \shortstack{\textbf{Element}\\\textbf{Recognition}}
& \shortstack{\textbf{Layout}\\\textbf{Understanding}}
& \shortstack{\textbf{Fine-grained}\\\textbf{Manipulation}}
& \textbf{Refusal}
& \textbf{Avg} \\
\midrule

JEDI-3B~\citep{xie2025scaling}
& 67.4 & 53.0 & 53.8 & 44.3 & 7.4 & 50.9 \\

Qwen3-VL-2B~\citep{bai2025qwen3}
& 61.7 & 45.8 & 54.2 & 39.6 & - & 45.9 \\

OS-ATLAS-7B~\citep{wu2025atlas}
& 44.1 & 29.4 & 35.2 & 16.8 & 7.4 & 27.7 \\

UGround-7B~\citep{gou2025uground}
& 51.3 & 40.3 & 43.5 & 24.8 & 0.0 & 36.4 \\

Aguvis-7B~\citep{xu2024aguvis}
& 55.9 & 41.2 & 43.9 & 28.2 & 0.0 & 38.7 \\

UI-TARS-7B~\citep{qin2025ui}
& 60.2 & 51.8 & 54.9 & 35.6 & 0.0 & 47.5 \\

UI-TARS-1.5-7B~\citep{qin2025ui}
& 36.8 & 62.7 & 62.2 & 50.8 & 0.0 & 52.8 \\

JEDI-7B~\citep{xie2025scaling}
& 65.9 & 55.5 & 57.7 & 46.9 & 7.4 & 54.1 \\

Qwen3-VL-8B~\citep{bai2025qwen3}
& 69.0 & 55.5 & 59.7 & 47.7 & - & 54.8 \\

GTA1-7B~\citep{gta1_2025}
& 42.1 & 65.7 & 62.7 & 56.1 & 0.0 & 55.1 \\

GUI-Owl-7B~\citep{mobileagentv3_2025}
& 64.8 & 63.6 & 61.3 & 41.0 & - & 55.9 \\

UI-Venus-7B~\citep{uivenus2025}
& 74.6 & 60.5 & 61.5 & 45.5 & - & 58.8 \\

Holo2-8B~\citep{hai2025holo2modelfamily}
& - & - & - & - & - & 62.2 \\

MolmoPoint-8B~\citep{clark2026molmopoint}
& - & - & - & - & - & 62.4 \\

COPC-8B~\citep{wang2026measure}
& - & - & - & - & - & 59.6 \\

\midrule

OpenCUA-32B~\citep{wang2026opencua}
& - & - & - & - & - & 59.6 \\

GUI-Owl-32B~\citep{mobileagentv3_2025}
& 67.0 & 64.5 & 67.2 & 45.6 & - & 58.0 \\

Qwen3-VL-32B~\citep{bai2025qwen3}
& 72.8 & 63.3 & 66.4 & 51.7 & - & 60.6 \\

GTA1-32B~\citep{gta1_2025}
& 63.2 & \textbf{78.4} & \textbf{73.3} & \textbf{65.2} & 0.0 & 65.2 \\

\midrule

GroundGUI-8B
& \textbf{77.0} & 68.5 & 66.4 & 57.2 & \textbf{22.2} & \textbf{65.4} \\

\bottomrule
\end{tabular}
\end{table*}

\begin{table*}[!tp]
\footnotesize
\centering
\caption{Performance comparison on \textbf{UI-Vision}  benchmark.
\textbf{Bold} values = best results.}
\label{tab:ui-vision}
\setlength{\tabcolsep}{16pt}
\renewcommand{\arraystretch}{0.90}

\begin{tabular}{l c c c c}
\toprule
\textbf{Model}
& \textbf{Basic}
& \textbf{Functional}
& \textbf{Spatial}
& \textbf{Avg} \\
\midrule

SeeClick~\citep{cheng2024seeclick}
& 9.4 & 4.7 & 2.1 & 5.4 \\

Qwen3-VL-2B~\citep{bai2025qwen3}
& 0.0 & 19.2 & 0.1 & 6.2 \\

OS-ATLAS-7B~\citep{wu2025atlas}
& 12.2 & 11.2 & 3.7 & 9.0 \\

UGround-V1-7B~\citep{gou2025uground}
& 15.4 & 17.1 & 6.3 & 12.9 \\

Qwen3-VL-8B~\citep{bai2025qwen3}
& 25.0 & 27.9 & 1.2 & 17.5 \\

UI-TARS-7B~\citep{qin2025ui}
& 20.1 & 24.3 & 8.4 & 17.6 \\

InfiGUI-G1-3B~\citep{liu2026infigui}
& 31.2 & 28.0 & 8.2 & 22.0 \\

UI-TARS-1.5-7B~\citep{qin2025ui}
& 28.8 & 27.5 & 10.7 & 22.3 \\

InfiGUI-G1-7B~\citep{liu2026infigui}
& 36.2 & 31.9 & 11.5 & 26.1 \\

UI-Venus-7B~\citep{uivenus2025}
& 36.1 & 32.8 & 11.9 & 26.5 \\

Phi-Ground~\citep{phiground2024}
& 36.8 & 37.1 & 7.6 & 27.2 \\

\midrule

UI-TARS-72B~\citep{qin2025ui}
& 31.4 & 30.5 & 14.7 & 25.5 \\

Qwen3-VL-32B~\citep{bai2025qwen3}
& 32.8 & 34.2 & 14.7 & 26.9 \\

UI-Venus-72B~\citep{uivenus2025}
& 45.6 & 42.3 & 23.7 & 36.8 \\

\midrule

GroundGUI-8B
& \textbf{46.3}
& \textbf{43.5}
& \textbf{26.3}
& \textbf{38.3} \\

\bottomrule
\end{tabular}
\end{table*}

We evaluate GroundGUI on 5 representative GUI grounding benchmarks:
ScreenSpot-V2 (SS-V2)~\citep{wu2025atlas},
MMBench-GUI L2 (MMB-GUI L2)~\citep{wang2025mmbench},
ScreenSpot-Pro (SS-Pro)~\citep{li2025screenspotpro},
OSWorld-G (OSW-G)~\citep{xie2025scaling},
and UI-Vision~\citep{nayak2025uivision}. ScreenSpot-V2 evaluates general-purpose GUI grounding across web, desktop, and mobile, covering text, icons, and interactive elements. MMBench-GUI L2 evaluates more complex instructions requiring stronger interface understanding beyond basic element matching across multiple platforms. ScreenSpot-Pro focuses on high-resolution professional software, challenging models to localize small targets and distinguish densely arranged, visually similar UI elements. OSWorld-G derives from realistic computer-use scenarios and emphasizes layout understanding, operation semantics, and fine-grained interaction reasoning. UI-Vision evaluates basic, functional, and spatial instructions, emphasizing reasoning over element semantics and relative spatial relationships. 
Baseline scores are taken from the corresponding publications.

\subsection{Comparison with Other Models}
As shown in Tables \ref{tab:screenspot_v2}--\ref{tab:ui-vision}, our GroundGUI-8B matches or outperforms all listed open-source 7B-scale baselines across all evaluated benchmarks, demonstrating strong GUI element grounding. Compared with the base model Qwen3-VL-8B~\citep{bai2025qwen3}, GroundGUI-8B achieves consistent performance gains: 3.5 points on ScreenSpot-V2, 5.5 on MMBench-GUI L2, 14.0 on ScreenSpot-Pro, 10.6 on OSWorld-G, and 20.8 on UI-Vision. These improvements are consistent with contributions from our ScreenRef data and D2RL, which jointly improve grounding across the evaluated benchmarks. Moreover, GroundGUI-8B surpasses Step-GUI-8B~\citep{yan2025step} by 1.3 points on ScreenSpot-Pro and 1.2 on MMBench-GUI L2. GroundGUI-8B delivers notably better results than Holo2-8B and MolmoPoint-8B on OSWorld-G, ScreenSpot-Pro, and ScreenSpot-V2.
Notably, Step-GUI and Holo2 keep their training datasets non-public, whereas we release our data, annotations, and code to facilitate further research. Beyond 8B counterparts, GroundGUI-8B also achieves stronger results than multiple 32B-scale models, including GUI-Owl-32B~\citep{mobileagentv3_2025}, ScaleCUA-32B~\citep{liu2026scalecua}, OpenCUA-32B~\citep{wang2026opencua}, and Qwen3-VL-32B. These results indicate that GroundGUI-8B attains strong
reported performance with far fewer parameters than these
models.

\subsection{Comparison with Other Datasets}
To directly compare with other open-source datasets, we train Qwen3-VL-8B on previous state-of-the-art grounding datasets, including ScaleCUA and GroundCUA. We explore the best training parameters for each dataset. As GroundCUA contains many images from UI-Vision, we omit the score on UI-Vision for fair comparison. As shown in Table \ref{tab:main-results-nouiv}, averaged over four benchmarks, the SFT-only model trained on ScreenRef  yields a 3.5-point gain over ScaleCUA and 2.7 points over GroundCUA. The results suggest that ScreenRef is more effective than these datasets on the evaluated benchmarks. On OSWorld-G, the ScreenRef model notably outperforms those trained on ScaleCUA and GroundCUA, by 3.2 and 5.3 points, respectively. A likely contributing factor is that ScreenRef provides refusal samples, which are absent in the other datasets. On ScreenSpot-Pro, the ScreenRef SFT model presents a larger advantage over ScaleCUA and GroundCUA, likely because our dataset comprises data across a broader diversity and higher annotation accuracy. 

\begin{table}[t]
\footnotesize
\centering
\caption{Qwen3-VL-8B fine-tuned on different grounding datasets under recommended settings for each.
UI-Vision is omitted due to source overlap with GroundCUA.
\textbf{Bold} values = best results.}
\label{tab:main-results-nouiv}

\setlength{\tabcolsep}{5pt}
\renewcommand{\arraystretch}{0.90}

\begin{tabular}{l cccc c}
\toprule
\textbf{Model}
& \textbf{SS-Pro}
& \textbf{OSW-G}
& \textbf{MMB-GUI L2}
& \textbf{SS-V2}
& \textbf{Average} \\
\midrule

Qwen3-VL-8B (reported)~\citep{zhao2026pointsguig}
& 49.9
& 54.8
& 81.3
& 91.7
& 69.4 \\

\quad + ScaleCUA SFT
& 52.8
& 61.7
& 83.6
& 93.6
& 72.9 \\

\quad + GroundCUA SFT
& 58.5
& 59.6
& \textbf{84.9}
& 91.9
& 73.7 \\

\midrule

\quad + \textsc{ScreenRef} SFT (ours)
& \textbf{62.4}
& \textbf{64.9}
& 84.5
& \textbf{93.9}
& \textbf{76.4} \\

\bottomrule
\end{tabular}
\end{table}

\subsection{Ablation Study}
We investigate the effectiveness of our method's key components (Table \ref{tab:rl-comparison}), where all post-training variants start from the same SFT checkpoint and use the same post-training data, and the single-stage variants (+OPSD and +RLOO) are each trained for 2 epochs. \textbf{OPSD:} Adding OPSD to SFT improves average score by 1.1 points, indicating that OPSD tailored for grounding improves the model via on-policy token-level supervision. \textbf{RLOO:} Applying RLOO after SFT yields a 0.9-point average gain. This result indicates that RLOO improves the SFT-only model by optimizing the policy toward target rewards. \textbf{D2RL:} D2RL surpasses SFT-only, SFT+OPSD, and SFT+RLOO on all benchmarks, supporting the method's effectiveness. Note that D2RL also outperforms RLOO$^{\dagger}$ by 0.8 points. These results  demonstrate that the benefit of D2RL is not due to extra exposure to the post-training data. D2RL's advantage comes from the stronger initial policy that OPSD provides.

\begin{table}[t]
\footnotesize
\centering
\caption{Ablation study of Distillation-to-Reinforcement Learning on all GUI grounding benchmarks. RLOO$^{\dagger} $ denotes a model first trained with SFT using 12K RL data and then optimized with RLOO. 
\textbf{Bold} values = best results.}
\renewcommand{\arraystretch}{0.90}
\setlength{\tabcolsep}{11pt}
\label{tab:rl-comparison}

\begin{tabular}{lcccccc}
\toprule
\textbf{Model}
& \textbf{SS-V2}
& \textbf{SS-Pro}
& \textbf{MMB-GUI L2}
& \textbf{UI-Vision}
& \textbf{OSW-G}
& \textbf{Average} \\
\midrule

SFT
& 93.9
& 62.4
& 84.5
& 34.3
& 64.9
& 68.0 \\

+ OPSD
& 95.0
& 62.6
& 85.8
& 37.0
& 65.1
& 69.1 \\

+ RLOO
& 95.0
& 63.4
& 85.8
& 36.5
& 64.0
& 68.9 \\
+ RLOO$^{\dagger}$
& 95.0
& 62.8
& 86.4
& 37.5
& 63.8
& 69.1 \\
\midrule

+ D2RL
& \textbf{95.2}
& \textbf{63.9}
& \textbf{86.8}
& \textbf{38.3}
& \textbf{65.4}
& \textbf{69.9} \\

\bottomrule
\end{tabular}
\end{table}

\section{Conclusion}
We introduce ScreenRef, a large-scale, high-quality GUI grounding dataset of 586K screenshots and 633K referring expressions across desktop, web, and mobile environments. Using ScreenRef, we construct a suite of real-world  instruction tasks for multimodal grounding. We also present GroundGUI-8B, trained via SFT followed by D2RL, which sequentially applies OPSD and RLOO. Across 5 benchmarks, GroundGUI substantially outperforms models trained on existing open-source datasets, while achieving performance competitive with or superior to leading models trained on closed-source data.

\subsection*{AI use statement}
In this work, we used generative AI models to generate synthetic datasets and to clean and reformat dataset content, as described in Section~\ref{sec:screenref}: Qwen3-VL-235B-A22B-Instruct generated grounding instructions from denoised metadata; Qwen3.8-27B verified instruction--target consistency; Qwen3.6-27B filtered non-tutorial video metadata; GPT-5.5 produced task segmentation, action annotations, and referring expressions from video frame pairs and transcripts; and Qwen3.7-Plus and Claude Opus 5 produced target coordinates with confidence gating. OmniParser V2 and YOLOv8 were used as non-generative detectors. We did not use generative AI tools to develop theoretical models or conceptual frameworks, formulate mathematical claims, write proofs, propose or refine hypotheses, design or evaluate the research methodology, translate, perform qualitative analysis, or interpret results. Additionally, we used generative AI tools to aid and polish the writing of this manuscript. All AI-generated data passed the automated verification stages described in Section~\ref{sec:screenref}, and all AI-assisted text was reviewed by the authors. We take responsibility for the final content of this work, including text, claims, and artifacts produced with the aid of generative AI.

\subsection*{Ethics statement}
ScreenRef is built from screenshots of live GUI environments and from frames of publicly available tutorial videos; no human subjects were recruited. The dataset used for D2RL post-training consists of samples from GroundCUA, excluding screenshots identical to those in the UI-Vision benchmark, and follows GroundCUA's MIT license. Screenshots of real interfaces may incidentally contain personal or account information, and we have not yet performed a systematic human privacy audit; the dataset documentation asks users to report any such content, and removal requests are actioned with a corrected revision. The code and model checkpoints are released under the Apache 2.0 license, and ScreenRef is released under a gated, non-commercial research-only license that prohibits redistribution, subject to source-platform terms. Improved GUI grounding can enable automated interaction with software, including unapproved automation on sensitive services; the released artifacts are intended for research on grounding, and their documentation states this intended use.

\subsection*{Reproducibility statement}
The dataset construction pipeline is described in Section~\ref{sec:screenref} and Figure~\ref{fig:screenref-pipeline}, including the models, thresholds (IoU 0.5, confidence $\tau=0.95$, at most 60 candidates per video), and verification stages used; D2RL is specified in Section~\ref{sec:opsd_rloo} with Equations~1--3 and Figure~\ref{fig:rl}; training data sizes, hardware, epochs, batch size, learning rate, and rollout count are given in Section~\ref{sec:impl}; benchmarks and the treatment of baseline scores are described in Section~\ref{sec:bench}; dataset statistics and examples are in Appendices~\ref{app:taxonomy-examples}--\ref{app:element-metadata}. The prompts, filtering rules, per-stage filtering statistics, and the full training configuration, including the position-aware weighting strength $\beta$, will be added to the code repository.

\bibliography{groundgui}
\bibliographystyle{groundgui}

\clearpage
\appendix
\section*{Appendix}

\section{Additional ScreenRef Examples}
\label{app:taxonomy-examples}

\begin{figure}[h]
\centering
\includegraphics[width=\textwidth]{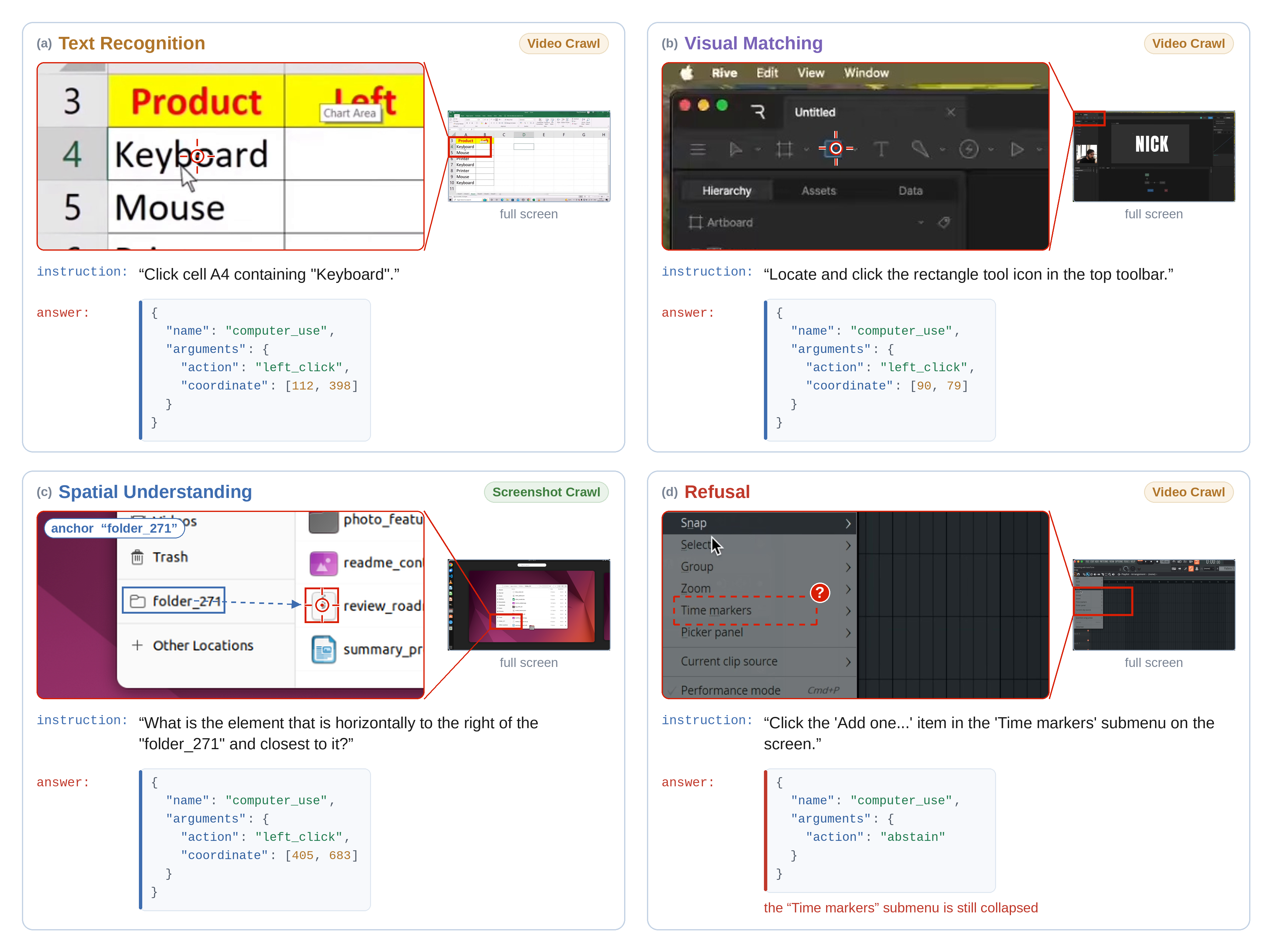}
\caption{Examples of the four ScreenRef task types in desktop environments.}
\label{fig:taxo-desktop}
\end{figure}

\begin{figure}[h]
\centering
\includegraphics[width=\textwidth]{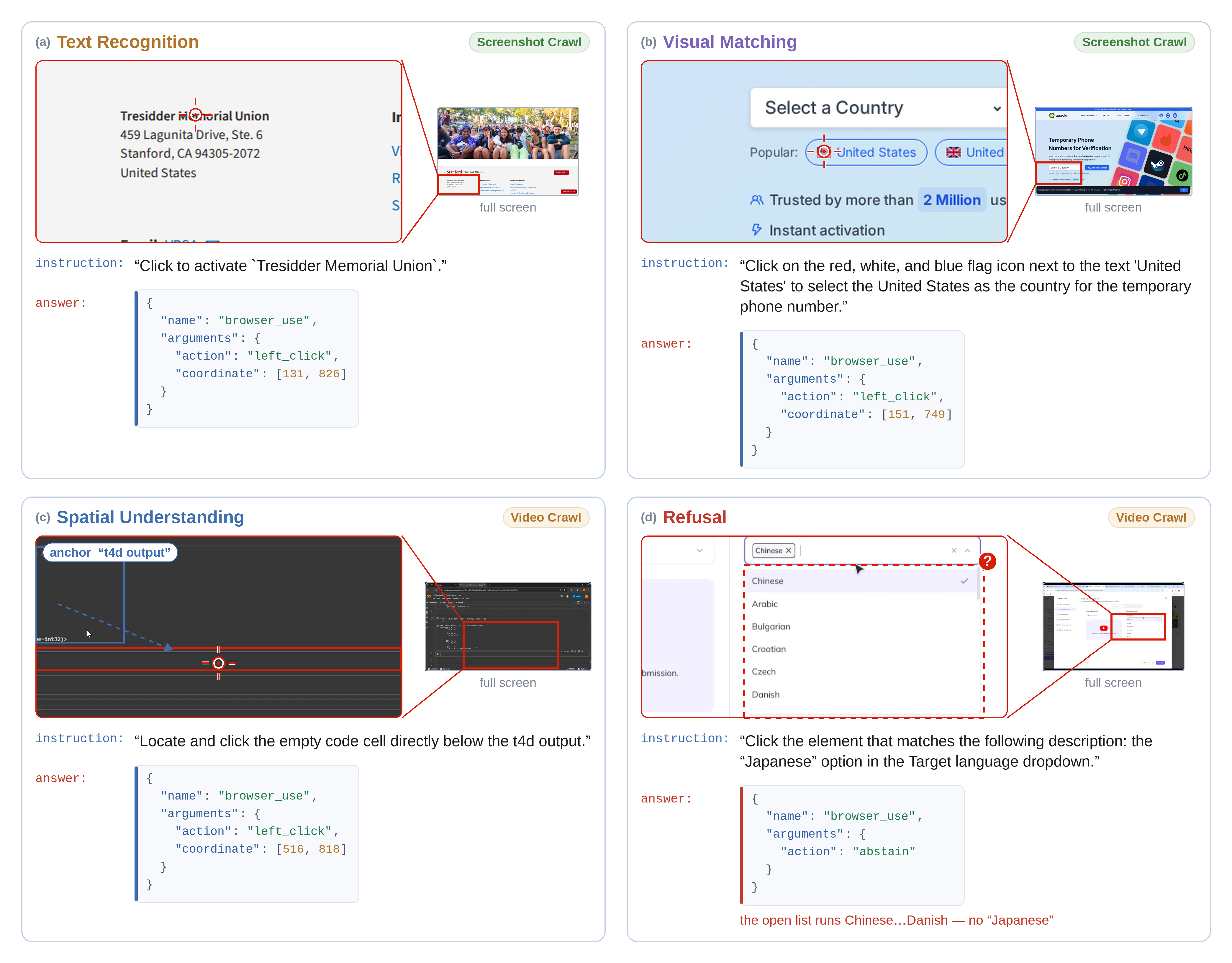}
\caption{Examples of the four ScreenRef task types in web environments.}
\label{fig:taxo-web}
\end{figure}

\begin{figure}[h]
\centering
\includegraphics[width=\textwidth]{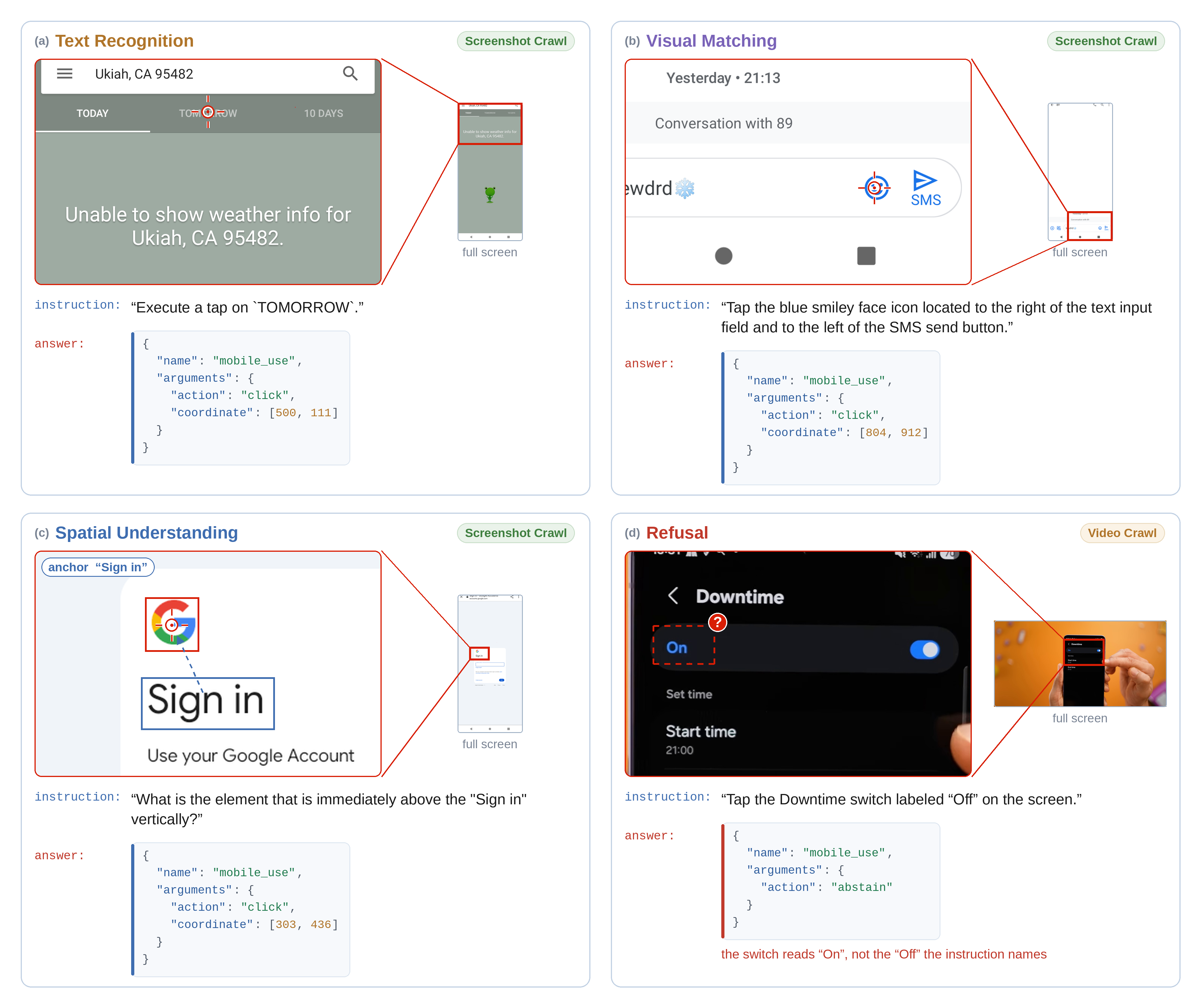}
\caption{Examples of the four ScreenRef task types in mobile environments.}
\label{fig:taxo-mobile}
\end{figure}

We provide additional qualitative examples of the four ScreenRef task types
across desktop, web, and mobile environments. Each answer is a tool call with an action and an integer coordinate pair normalized to $[0,1000]$ along the image width and height; refusal samples use an abstain action (Figures~\ref{fig:taxo-desktop}--\ref{fig:taxo-mobile}).

\section{Additional Details for Dataset Comparison} 
\label{app:dataset-comparison}

Figures for prior datasets are reported as published rather than recomputed here. For Images and
Res Range we align with the comparison in Table~1 of \citet{feizi2026grounding}, the paper that
introduces GroundCUA and is itself a row here, so that these columns rely on consistent measurements across datasets rather than totals produced under different protocols; where that
comparison does not cover a dataset we use the figure reported in its own paper. They differ in
places from a dataset's own stated total, which may reflect different filtering. Instr.\ is
each paper's own reported count, left blank where none is given rather than derived from element
counts. Platform follows each paper's account of its data, and where a reported figure covers
only part of a dataset the marks follow that part.

Refusal marks grounding samples whose referred target is absent from the screenshot; a
task-level infeasible action, which several agent datasets provide, is not counted. A pipeline
that derives every instruction from an element present in the screenshot yields no such sample,
which is the case for every prior dataset here except JEDI, which pairs instructions with
unrelated screenshots. Figures for \textsc{ScreenRef} are measured on the final dataset.

\section{Additional ScreenRef Statistics} \label{app:screenref-statistics} We provide additional statistics on the distribution of grounding capabilities in ScreenRef. Table~\ref{tab:data-capability} summarizes the number of grounding instances across desktop, web, and mobile platforms for the four capability categories: Text Recognition, Visual Matching, Spatial Understanding, and Refusal. \begin{table}[t] \footnotesize \centering \caption{Distribution of grounding capabilities across platforms.} \label{tab:data-capability} \begin{tabular}{lccccc} \toprule Platform & Text & Visual & Spatial & Refusal & Total \\ \midrule Desktop & 128,018 & 61,951 & 63,291 & 13,912 & 267,172 \\ Web & 167,115 & 62,294 & 67,799 & 6,582 & 303,790 \\ Mobile & 24,160 & 12,526 & 23,740 & 1,981 & 62,407 \\ \midrule \textbf{Total} & \textbf{319,293} & \textbf{136,771} & \textbf{154,830} & \textbf{22,475} & \textbf{633,369} \\ \bottomrule \end{tabular} \end{table} Text Recognition constitutes the largest portion of ScreenRef, accounting for 50.4\% of all grounding instances, followed by Spatial Understanding (24.4\%), Visual Matching (21.6\%), and Refusal (3.5\%). The capability distribution also varies across platforms. Text Recognition represents the largest category on both web (55.0\%) and desktop (47.9\%) interfaces, whereas the mobile subset is more balanced between Text Recognition (38.7\%) and Spatial Understanding (38.0\%). Desktop interfaces contain the highest proportion of Visual Matching instances (23.2\%) and Refusal instances (5.2\%), reflecting the greater prevalence of visually identified controls and absent-target scenarios in desktop environments.

\newpage
\vspace*{8pt}
\section{Element Metadata from Live Screenshot Crawling}
\label{app:element-metadata}

\begin{figure}[h]
\centering
\includegraphics[width=\textwidth]{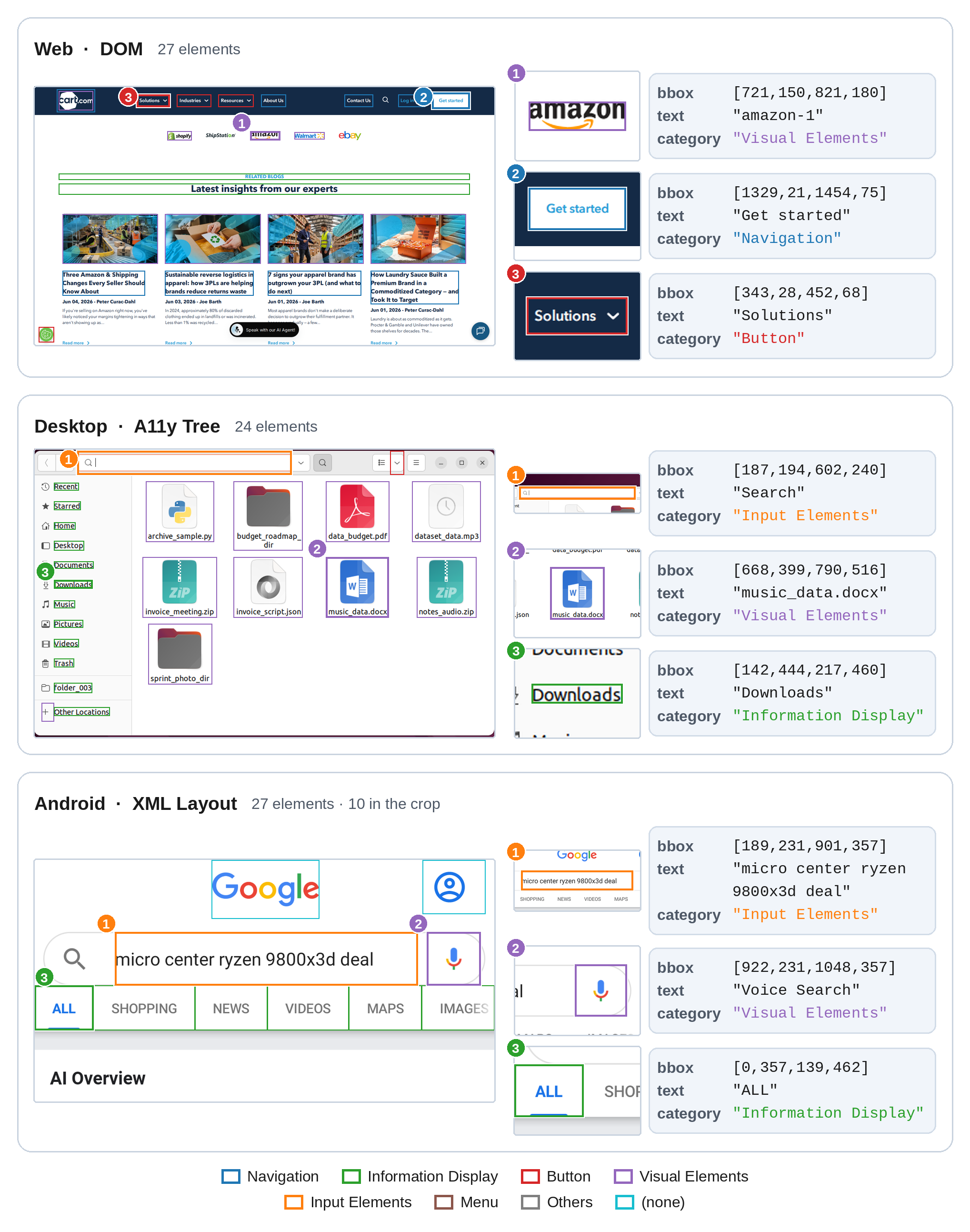}
\caption{Examples of element metadata from live screenshot crawling. Left: a real screenshot,
cropped to the region containing metadata, with all element boxes in this region that survive
filtering, colored by category (``(none)'' marks elements without a category). Right: three
numbered elements from the same screenshot, zoomed in, with their stored records (bounding box,
text, and category).}
\label{fig:element-metadata}
\end{figure}

Part of ScreenRef is built from live screenshot crawling, where we capture screenshots from real
web pages, desktop applications, and Android applications. Alongside each screenshot, we record
element metadata from the platform's native interface: the DOM on web pages, the accessibility
tree on desktop, and the XML layout on Android.
Each element record contains a bounding box in the pixel coordinates of the screenshot, a text
string, and a functional category such as Button, Input Elements, or Visual Elements.
The text is the element's visible text or, if the element displays none, its accessibility name;
the category may be absent for some Android elements.
Every record contains non-empty text, so the metadata covers the elements that the interface
exposes with a text or an accessibility name, rather than every visual object on the screen.
Before the metadata is used to generate grounding instructions, it is filtered to remove or refine
unreliable screenshots and boxes. Figure~\ref{fig:element-metadata} shows real examples of the
filtered metadata.

\end{document}